\documentclass[letterpaper,10pt,conference]{ieeeconf}

\IEEEoverridecommandlockouts
\usepackage[table,xcdraw,dvipsnames]{xcolor}
\usepackage{soul}
\setstcolor{red}

\usepackage{graphicx}
\usepackage{epsfig} % for postscript graphics files
\usepackage{amsmath} % assumes amsmath package installed

\usepackage{amssymb}  % assumes amsmath package installed
\usepackage{amsthm}
\usepackage{wasysym}
\usepackage[mathcal]{euscript}
\usepackage{bbm}
\usepackage{color}
\usepackage{balance} % used for preventing ragged columns on the last page.

\usepackage{algorithm}
\usepackage[noend]{algpseudocode}
\algrenewcommand\algorithmicindent{1.0em}% 
\usepackage{hyperref}
\usepackage{etoolbox}
\usepackage[table,xcdraw,dvipsnames]{xcolor}
\usepackage{multirow}
\usepackage{mathtools}
\usepackage{outlines}
\let\labelindent\relax
\usepackage{enumitem}
\setenumerate[1]{label=\arabic*.}
\setenumerate[2]{label=\alph*.}

\newcommand{\pcnote}[1]%
    {\textcolor{cyan}{\textbf{PC: #1}}}
    
\newcommand{\rgnote}[1]%
    {\textcolor{orange}{\textbf{RG: #1}}}
    
\newcommand{\acnote}[1]%
    {\textcolor{violet}{\textbf{AC: #1}}}

\newcommand{\manote}[1]%
    {\textcolor{green}{\textbf{MA: #1}}}
    
\newcommand{\msnote}[1]%
    {\textcolor{blue}{\textbf{MS: #1}}}
    
\newcommand{\jbnote}[1]%
    {\textcolor{red}{\textbf{JB: #1}}}

\newcommand{\edit}[1]{{\leavevmode\color{black}#1}}

\usepackage{subfigure}
\graphicspath{ {figures/} {figures/estimator/}}
\usepackage[acronyms, shortcuts]{glossaries}
\glsdisablehyper

\newacronym{nerf}{NeRF}{Neural Radiance Field}
\newacronym{sdf}{SDF}{Signed Distance Field}
\newacronym{nlp}{NLP}{nonlinear program}
\newacronym{mle}{MLE}{maximum likelihood estimation}

\renewcommand{\baselinestretch}{0.975} % this is bad practice and should not be used ;)

\begin{document}
\title{\LARGE \bf
Vision-Only Robot Navigation in a Neural Radiance World %*\\
}

\author{Michal Adamkiewicz,*$^1$ Timothy Chen,*$^2$ Adam Caccavale,$^3$ Rachel Gardner,$^1$ Preston Culbertson,$^3$\\ Jeannette Bohg,$^1$ Mac Schwager$^2$%

\thanks{*These authors contributed equally.}%
\thanks{$^1$Department of Computer Science, Stanford University, Stanford, CA, 94305, USA {\texttt\small \{mikadam, rachel0, bohg\}@cs.stanford.edu}}
\thanks{$^2$Department of Aeronautics and Astronautics, Stanford University, Stanford, CA 94305, USA, {\texttt\small \{chengine, schwager\}@stanford.edu}}
\thanks{$^3$Department of Mechanical Engineering, Stanford University, Stanford, CA 94305, USA, {\texttt\small \{awc11, pculbertson\}@stanford.edu}}
\thanks{This work was supported in part by NSF NRI grant 1830402, ONR grant N00014-18-1-2830, Siemens and the Stanford Data Science Initiative. P. Culbertson was supported by a NASA Space Technology Research Fellowship.}
% \footnotetext{* denotes equal contribution} 
}

% \and
% \and

\maketitle

\begin{abstract}
 \acp{nerf} have recently emerged as a powerful paradigm for the representation of natural, complex 3D scenes. \acp{nerf} represent continuous volumetric density and RGB values in a neural network, and generate photo-realistic images from unseen camera viewpoints through ray tracing.  We propose an algorithm for navigating a robot through a 3D environment represented as a \ac{nerf} using only an on-board RGB camera for localization.  We assume the \ac{nerf} for the scene has been pre-trained offline, and the robot's objective is to navigate through unoccupied space in the \ac{nerf} to reach a goal pose.  We introduce a trajectory optimization algorithm that avoids collisions with high-density regions in the \ac{nerf} based on a discrete time version of differential flatness that is amenable to constraining the robot's full pose and control inputs.  We also introduce an optimization based filtering method to estimate 6DoF pose and velocities for the robot in the \ac{nerf} given only an onboard RGB camera.  We combine the trajectory planner with the pose filter in an online replanning loop to give a vision-based robot navigation pipeline.  We present simulation results with a quadrotor robot navigating through a jungle gym environment, the inside of a church, and Stonehenge using only an RGB camera. We also demonstrate an omnidirectional ground robot navigating through the church, requiring it to reorient to fit through a narrow gap. Videos of this work can be found at \href{https://mikh3x4.github.io/nerf-navigation/}{mikh3x4.github.io/nerf-navigation/}. %We also show a ground robot navigating through the inside of a house.  
\end{abstract}

% \begin{IEEEkeywords}
% TO DO
% \end{IEEEkeywords}
\glsresetall % reset acronym counts so they're defined in the paper body.

\section{Introduction}
\label{Sec:Introduction}

Planning and executing a trajectory with on-board sensors is a fundamental building block of many robotic applications, from manipulation to autonomous driving or drone flight.  Robot navigation methods depend on properties of the underlying environment representation, whether it is a voxel grid, a point cloud, a mesh model, or a \ac{sdf}. Recently there has been an explosion of interest in a deep-learned geometric representation called \acp{nerf} due to their ability to compactly encode detailed 3D geometry and color \cite{mildenhall2020nerf}.  \acp{nerf} take a collection of camera images and train a neural network to give a function relating each 3D point in space with a density and a vector of RGB values (called a ``radiance'').  This representation can then generate synthetic photo-realistic images through a differentiable ray tracing algorithm.  In this paper, we propose a navigation pipeline for a robot given a pre-trained \ac{nerf} of its environment.  We use the density of the \ac{nerf} to plan dynamically feasible, collision-free trajectories for a differentially flat robot model. We also build a filter to estimate the dynamic state of the robot given an on-board RGB image, using the image synthesis capabilities of the \ac{nerf}.  

% We combine the trajectory planner and the filter in a receding horizon loop to provide a full navigation pipeline for a robot to dynamically maneuver through an environment using only an RGB camera for feedback. Existing vision-only navigation systems such as~\cite{2020_TRO_DeepDrone} have also advocated for a modularization of learned perception and control and achieved impressive results. Their perception system aims to generalize over variations of drone race tracks and requires very specific training data and labels. We focus on \acp{nerf} as a geometric environment representation that enables any robot, e.g. drones or ground robots, to navigate through it.

We combine the trajectory planner and the filter in a receding horizon loop to provide a full navigation pipeline for a robot to dynamically maneuver through an environment using only an RGB camera for feedback. \edit{While some existing vision-only navigation systems \cite{loquercio2021} have seen recent success with end-to-end approaches, others} such as~\cite{2020_TRO_DeepDrone} have advocated for a modularization of learned perception and control and achieved impressive results. Their perception system aims to generalize over variations of drone race tracks and requires very specific training data and labels. We take a similar approach, and focus on \acp{nerf} as a geometric environment representation that enables any robot, e.g. drones or ground robots, to navigate through it.

\begin{figure}[t]
    \centering
    \includegraphics[width=\linewidth]{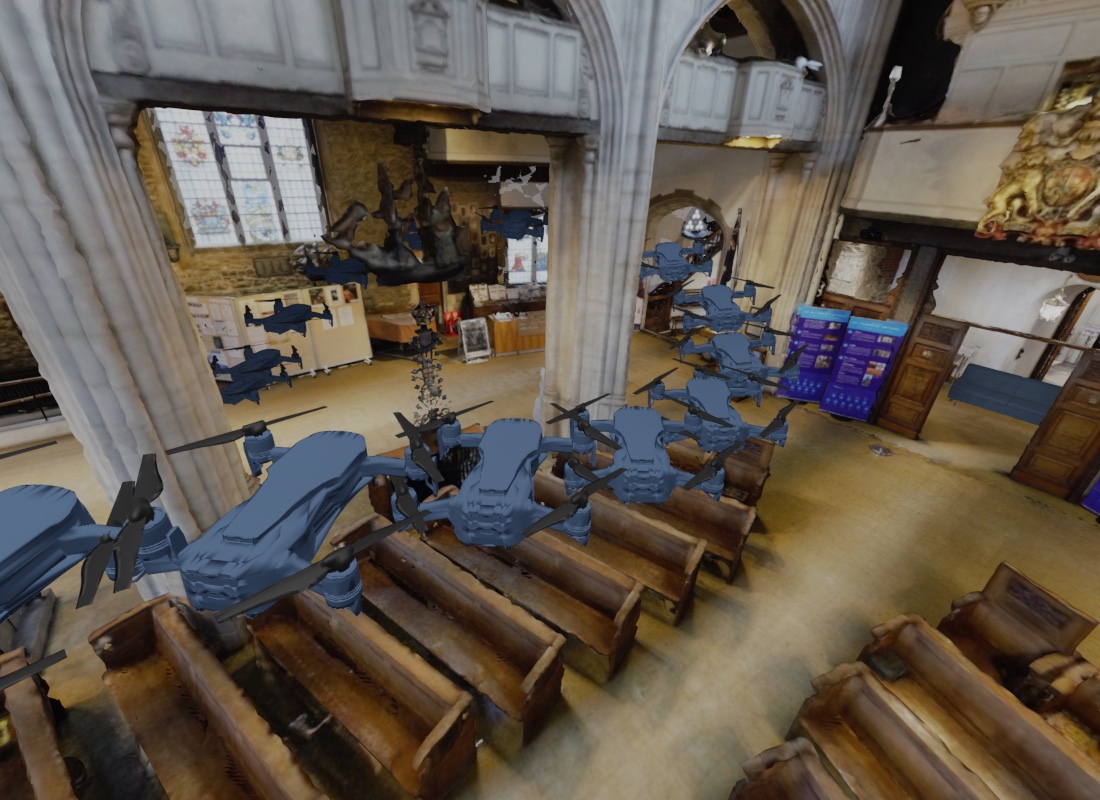}
    \caption{A drone navigating through the interior space of a church using a monocular camera. The environment is modeled as a \acf{nerf}, a deep-learned geometry representation. % which has seen an explosion of interest in the computer vision community for novel view synthesis of photo-realistic images. 
    The trajectory, which is optimized to minimize a \ac{nerf}-based collision metric, can be continually replanned as the drone updates its state estimate based on captured images.}
    %\vspace{-1em}
    \label{fig:teaser_image}
\end{figure}

%\subsection{Motivation}
% \paragraph*{Motivation}

\acp{nerf} present a range of potential advantages as an environment representation for robots.  Unlike voxel models, point clouds, or mesh models, they are trained directly on dense photographic images without the traditional feature extraction, matching, and alignment pipeline \cite{sucarIMAPImplicitMapping2021, wang2021nerfmm}.  They inherently represent the geometry as a continuous density field, and so they are well-suited to robot motion planning and trajectory optimization. They can also produce photo-realistic synthetic images, giving a mechanism for a robot to ``hallucinate" what it would expect to see if it were to take different actions.  

\edit{Extensions of \acp{nerf} have been developed to handle transparent objects \cite{2021nerfGTO}, segment and recompose objects in a scene \cite{yuan2021star, niemeyerGIRAFFERepresentingScenes2021, ostNeuralSceneGraphs2021}, and render moving and deformable objects \cite{pumarolaDNeRFNeuralRadiance2020}, including humans \cite{pumarolaDNeRFNeuralRadiance2020} and human faces \cite{gafni2021dynamic}.  Variants of \acp{nerf} can also incorporate a prior over objects or scenes, in order to quickly adapt to new environments with a handful of images \cite{tancikLearnedInitializationsOptimizing2021, yu2021pixelnerf}. While the standard \ac{nerf} training and image rendering pipeline is slow, recent developments have accelerated image synthesis from a \ac{nerf} to 200 frames per second on a GPU \cite{garbin2021fastnerf}, fast enough for use in a real-time robot control loop.} \edit{We envision a future where all these improvements could be leveraged to create a fully \ac{nerf} based environment representation that combines complex geometry, semantic understanding, and real-time performance.} However, for robots to harness the advantages of the \ac{nerf} representation for navigation, a trajectory planner and pose filter designed to work specifically with the \ac{nerf} machinery are needed.

\edit{We address this need in this paper by proposing:
\begin{itemize}
    \item a new trajectory planning method, based on differential flatness, that plans full, dynamically feasible trajectories to avoid collisions with a \ac{nerf} environment,
    \item an optimization-based filter to obtain a posterior estimate of the full dynamic state, balancing a dynamics prediction loss and a \ac{nerf}-based photometric loss, and 
    \item an online replanning controller that combines both of the above algorithms in feedback to robustly navigate a robot through its environment using a \ac{nerf} model.
\end{itemize}}

We demonstrate results in a variety of high fidelity simulation environments, and perform ablation studies to showcase the advantages provided by each part of our navigation framework. We run our navigation pipeline with custom-trained \ac{nerf} models of a playground, a church, and Stonehenge.  We then evaluate the performance of our trajectory planner and pose estimator on the underlying ground truth mesh models, not the trained \ac{nerf} models, thereby demonstrating robustness to model mismatch between the real-world scene and the trained \ac{nerf}.

\section{Related Work}
\label{Sec:Related}

%\begin{itemize}
%    \item Original nerf paper with one sentence what they are (just copy from abstract)
%    \item Describe their impact
%    \item Include work on localization of robot within \ac{nerf}
%    \item How is your work different from those prior works?
%    \item Are there any works doing trajectory optimization in \acp{nerf}? I don't think so, so maybe talk about Trajectory opt using other representations like SDFs etc
%\end{itemize}

\subsection{Neural implicit representations}

% \begin{figure}[t]
%     \centering
%     \includegraphics[width=0.35\textwidth]{figures/nerf_explainer.pdf}
%     \caption{Schematic of the differentiable rendering process for a \ac{nerf}. \textbf{Top:} For each pixel in the synthetic image, a ray is cast into the environment. At sample points along the ray, the \ac{nerf} is queried for its density (shown by the circle size) and radiance (shown by the circle fill color). \textbf{Bottom:} The sampled densities are converted to termination probabilities. The color of the pixel is then calculated as the expected color given these termination probabilities (e.g., this pixel is rendered as gray).}
%     %\vspace{-1.25em}
%     \label{fig:nerf_explainer}
% \end{figure}

\begin{outline}

Neural implicit representations use a neural network to represent the geometry (and sometimes the color and texture) of a complex 3D scene.  Generally, neural implicit representations take a labeled data set and learn a function of the form \(f_\theta(p) = \sigma\), where \(f\) is a neural network parameterized by the weights \(\theta\), \(p\) is a low-dimensional query point such as an \((x, y, z)\) coordinate, and \(\sigma\) is some (usually scalar) quantity of interest. Aside from \ac{nerf}s, there are several other approaches to implicit representations including learned \acp{sdf} \cite{michalkiewiczImplicitSurfaceRepresentations2019, parkDeepSDFLearningContinuous2019, sitzmannImplicitNeuralRepresentations2020b} and Occupancy Networks \cite{meschederOccupancyNetworksLearning2019}.  %\acf{sdf} representations are another common implicit representation, though one only recently to which neural networks have been applied \cite{parkDeepSDFLearningContinuous2019, others}. 

However, there currently exists little work studying how to leverage \acp{nerf} for applications beyond novel view synthesis. Recent work \cite{sucarIMAPImplicitMapping2021} has treated the problem of mapping and online \ac{nerf} construction from visual data; the authors demonstrate competitive accuracy with traditional SLAM pipelines, and realtime performance. This work's state estimator builds on \cite{yen2020inerf}, which presents a method for single-image camera pose estimation using a pre-trained \ac{nerf} representation of the environment. The method we present here for state estimation also uses 
\ac{mle}, but we instead treat the problem as recursive Bayesian estimation, which incorporates system dynamics and must propagate uncertainty between timesteps. % A handle of papers do mapping \cite{sucarIMAPImplicitMapping2021} and localization \cite{yen-chenINeRFInvertingNeural2021}, but to our knowledge, our work is the first to tackle trajectory estimation using a \ac{nerf}.    

\end{outline}

\subsection{Trajectory optimization}
Optimal control remains a fundamental tool in robotic motion planning. Of particular interest is the problem of trajectory optimization \cite{kelly2017}, which seeks a system trajectory $\mathbf{x}(t)$ and open-loop inputs $\mathbf{u}(t)$ that optimize a control objective, subject to state and input constraints. While there exists a vast literature on trajectory optimization for robot motion planning \cite{betts2010}, our discussion here will focus specifically on collision avoidance in trajectory optimization, which remains unstudied for environments represented as \acp{nerf}.

% A typical approach to trajectory optimization, direct transcription, poses the problem as a \ac{nlp}, and solves for a finite set of states and controls, enforcing the system dynamics with equality constraints. Collision constraints are typically included explicitly as inequality constraints on the system states. For polygonal or circular objects, the constraints may be expressed using linear inequalities or $\ell_p$-norms (as in \cite{howell2019}); modeling general geometry is more difficult. 

One approach to model an environment is as an \ac{sdf}, which represents obstacles as the zero-level set of a nonlinear function $d(\mathbf{x}),$ which takes \edit{negative values inside the obstacle, positive values outside the obstacle}, and has magnitude equal to the distance between $\mathbf{x}$ and the obstacle boundary. Collision avoidance is typically imposed as a constraint in the trajectory optimization, requiring the \ac{sdf} for all obstacles to be \edit{non-negative} at all points on the robot body along the trajectory. \edit{This formalism has received particular interest as a map representation following the success of KinectFusion \cite{newcombe2011}, which constructs truncated \acp{sdf} using RGB-D data. Works such as \cite{oleynikova2017} and \cite{han2019} present methods for incrementally constructing \ac{sdf}-like map representations and using them for online motion planning. }

% This approach is common in robotic trajectory optimization \cite{howell2019,bonalli2019}, especially when performing contact-implicit trajectory optimization \cite{posa2014}, which seeks to reason about contact geometry when optimizing contact sequences for manipulation or locomotion. 
 
Perhaps closest to this work's trajectory optimizer is CHOMP \cite{ratliff2009, mainprice2016}, \edit{a family of gradient-based methods} which optimizes a finite sequence of poses, with an objective which encourages the trajectory to be smooth and to avoid collision. Specifically, CHOMP represents obstacle geometry by pre-computing each obstacle's \ac{sdf} on a finite grid, and approximates \ac{sdf} gradients using finite differences or interpolation. \edit{In \cite{zhou2021}, the authors present a similar gradient-based method which optimizes quadrotor position trajectories to minimize a perception-aware collision metric based on an SDF-like map.} In contrast, the \ac{nerf} geometry representation used here is continuous in itself, and of arbitrary resolution, with continuous gradients that can be efficiently computed using automatic differentiation. Further, our method generates trajectories that are constrained to be dynamically feasible rather than imposing the system dynamics via a cost.

\section{Problem Formulation}
\label{Sec:ProbForm}

This paper proposes a method for navigating a robot through an environment represented by a \ac{nerf}. A \ac{nerf} ($N:\mathbb{R}^3 \times \mathbb{R}^2 \mapsto \mathbb{R}^3 \times \mathbb{R}_+$) maps a 3D location $\mathbf{p} = (x, y, z)$ and view direction $(\theta, \phi)$ to an emitted color $\mathbf{c}=(r,g,b)$ and scalar density $\rho.$ For notational convenience, we define $\rho(\mathbf{p})$ as the density output of the \ac{nerf} evaluated at position $\mathbf{p}$ (note $\rho$ depends only on position). Similarly, we define $C_i: \text{SE}(3) \mapsto \mathbb{R}^3$ as the expected color of pixel $i$ when rendering the \ac{nerf} from the camera pose $T \in \text{SE}(3)$, where $\text{SE}$ denotes the special Euclidean group.

In this paper, we consider the problem of a mobile robot, equipped only with a monocular camera, which seeks to navigate an environment. Specifically, the robot seeks to plan and track a collision-free path from its initial state $\mathbf{x}_0$ to a goal state $\mathbf{x}_f$. The robot has access to a \ac{nerf} representation of the environment which it can use for both planning (i.e., for evaluating the probability of collision for a given trajectory) and localization. 

We approximate the robot body using a finite collection of points $\mathcal{B}$ at which collision is checked. Typically this will be a 3d grid of points representing the robot's bounding box, however it can also be an arbitrarily complex model. However, it is not obvious how the \ac{nerf} density at a point relates to its occupancy. Specifically, the \ac{nerf} density represents the differential probability of a given spatial point stopping a ray of light \cite{mildenhall2020nerf}. We assume the probability of terminating a light ray is a strong proxy for the probability of terminating a mass particle. Thus, the collision probability at time $t$ is given by
\begin{align}
    p^\text{coll}_t = P\left( \bigcup_{\; \mathbf{b}_t \in \mathcal{B}} \mathbf{b}_t \in \mathcal{X}_\text{coll}\right) \geq \sum_{\mathbf{b}_t \in \mathcal{B}} \rho(\mathbf{b}_t) \;  s(\mathbf{b}_t),
    \label{eqn:collision_prob}
\end{align}
where $\mathcal{X}_\text{coll}$ denotes the collision set, $s(\mathbf{b}_t)$ is the distance traveled by a body-fixed point $\mathbf{b}$ in timestep $t$, and the bound follows from Boole's inequality. In this work, we include the collision probability as a cost to be minimized during trajectory optimization; an alternative approach would be to impose a chance constraint on the optimization, which would require more sophisticated optimization techniques.

%  Ourinclude this approximate collision probability as a weighted cost. Alternatively one could impose a chance constraint, but this would necessitate more sophisticated optimization techniques. 

%When planning a trajectory from $\mathbf{x}_0$ to $\mathbf{x}_f$, it is essential that the robot avoids collision with the environment. However, since the \ac{nerf} only provides an obstacle density (i.e., a differential occupancy probability) at each point, it is not obvious how to determine whether a given robot trajectory is in collision. In this work, we appoximate the robot body using a finite collection of points $\mathcal{B}$ at which collision is checked. Thus, for a set of obstacles given by $\mathcal{X}_\text{coll},$ the collision probability at time $t$ is given by
% \begin{align*}
%     p^\text{coll}_t &= P\left( \bigcup_{\; \mathbf{b}_t \in \mathcal{B}} \mathbf{b}_t \in \mathcal{X}_\text{coll}\right),\\
%     &\geq \sum_{\mathbf{b}_t \in \mathcal{B}} \rho(\mathbf{b}_t) \;  s(
%     \mathbf{b}_t),
% \end{align*}
%where $s(\mathbf{b}_t)$ is the distance traveled by a body-fixed point $\mathbf{b}$ in timestep $t$, and the second inequality follows from Boole's inequality. Since the density $\rho$ is a differential collision probability, we scale it by the arc length $s$ to evaluate the collision probability for each point. Thus, our trajectory optimization problem is to minimize an objective for the trajectory, subject to the collision probability $p^\text{coll}$ being below a specified threshold $\alpha$.
%\pcnote{TODO: reword}

Given a Gaussian estimate of its current state, $\mathcal{N}(\mu_t, \Sigma_t)$, the robot plans a series of waypoints that avoid regions of high density in the \ac{nerf}. After taking a control action, the robot receives an image of the environment, and updates its belief about its current state. Finally, the robot replans the trajectory using the latest estimate as the first state.

% Using the principle of differential flatness \cite{nieuwstadt1998}, only four state elements are optimized at each waypoint $\sigma = [x, y, z, \psi]$, where $\psi$ is the yaw angle in the world frame. The rest of the state elements, as well as the control actions, are calculated from this state subset as well as the adjacent waypoints.

\begin{figure}[t]
    \centering
    \includegraphics[clip, trim=0cm 0cm 0cm 0cm, width=0.45\textwidth]{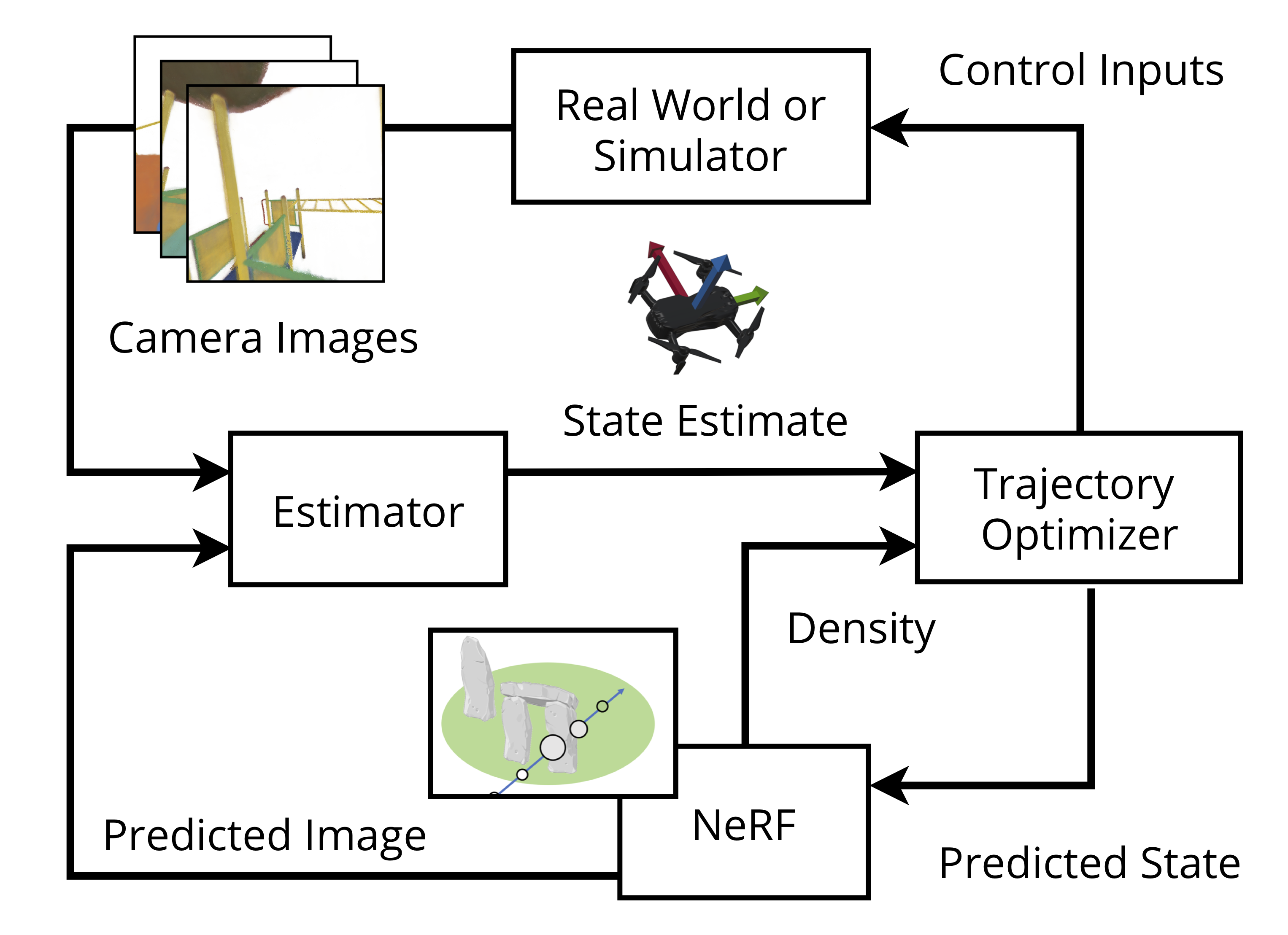}
    \caption{Block diagram of the proposed pipeline. Our method consists of a trajectory optimizer and state estimator which use a \ac{nerf} representation of the environment for planning and localization. At each timestep, the planner optimizes a trajectory from the current mean state estimate which minimizes a \ac{nerf}-based collision metric. The robot then applies the first control action of this trajectory, and receives a noisy image from its onboard camera. Finally, the state estimator, using the \ac{nerf} as a nonlinear measurement model, uses this image to generate a posterior belief over the new state.}
    %\vspace{-1.25em}
    \label{fig:system_block_diagram}
\end{figure}
\section{Trajectory Planning in a \ac{nerf}}
\label{Sec:TrajectoryPlanning}

This paper addresses the unique challenges that prevent common trajectory planning methods from working with \ac{nerf} environment representations. Querying a \ac{nerf} at a point in space gives a density, not an absolute occupancy, which prevents the use of hard constraints and instead suggests a method that seeks to minimize the integrated density over the volume of the robot.

\subsection{Differential flatness}
To speed up planning, our system leverages ``differential flatness,'' a particular property of some dynamical systems which allows their inputs and states to be represented using a (smaller) set of ``flat outputs,'' and their derivatives. Notably, quadrotors are known to be differentially flat, with their position and yaw angle as flat outputs \cite{mellinger2011}. 
% A dynamical system is differentially flat \cite{nieuwstadt1998} if it admits a set of ``flat outputs'' such that all inputs and states of the system can be described using algebraic functions of the outputs and their derivatives. This property is often desirable, since trajectories for differentially flat systems may be planned in the (lower-dimensional) output space, and then mapped to a set of desired inputs. Notably, quadrotors are known to be differentially flat, with their position and yaw angle as flat outputs. \cite{mellinger2012, mellinger2011}. 

% Differential flatness, as it is most commonly used, is incompatible with a \ac{nerf} environment representation due to the required hard constrains the are part of its core optimization formulation. Instead, we propose a discrete time variation of the method where the trajectory is represented as a series of waypoints to be optimized, and using finite differencing in place of analytical derivatives, allowing us to achieve a dynamically feasible trajectory without hard constraints. This is contrast to the usual method where a quadradic program is solved to define fixed waypoints, which are then fit with a smooth spline. While our discrete time differential flatness method will work for any system where the control actions and full state can be analytically derived from a subset of the states, we explicitly derive the equations for quadrotors.

Traditional planning pipelines for differentially flat systems \cite{mellinger2012, nieuwstadt1998a} seek polynomial trajectories for the flat outputs which minimize an objective functional (such as snap or jerk) subject to waypoint constraints. This problem can be expressed as a quadratic program, which can be solved efficiently. Collision avoidance can also be included in this formulation, but in order for the problem to remain convex, the designer must hard-code decisions about how obstacles will be passed. 

Our approach differs from the traditional pipeline since we do not describe the obstacles in closed form (e.g., as polytopes), but instead represent them implicitly using the \ac{nerf} density. 
\edit{Additionally, while prior methods only optimize the trajectory between static, hand-designed waypoints, our method uses a denser set of waypoints whose location can be optimized directly}. Because our trajectory optimization is fundamentally nonconvex, we instead perform our optimization using first-order methods (in particular, the Adam optimizer) with gradients computed efficiently using automatic differentiation. Our decision variables thus are a set of flat output waypoints that we optimize to minimize a combined objective of collision probability and control effort. One advantage of our approach is that the cost \edit{ can be an arbitrary differentiable functional of the trajectory or robot state}; further, our planned trajectory can be naturally combined with differential flatness-based feedback controllers for low-level tracking. \edit{Note that while this paper uses quadrotors as an example,} this property is known to hold for numerous other vehicle types, such as omnidirectional or differential drive ground robots. 

%two common systems: a quadrotor and an omnidirecitonal ground robot/differicanl drive robot

\subsection{Optimization formulation}

\begin{figure}[t]
\centering
\subfigure[][]{%
    \includegraphics[clip, trim=0cm 0cm 0cm 0cm, width=0.425\linewidth]{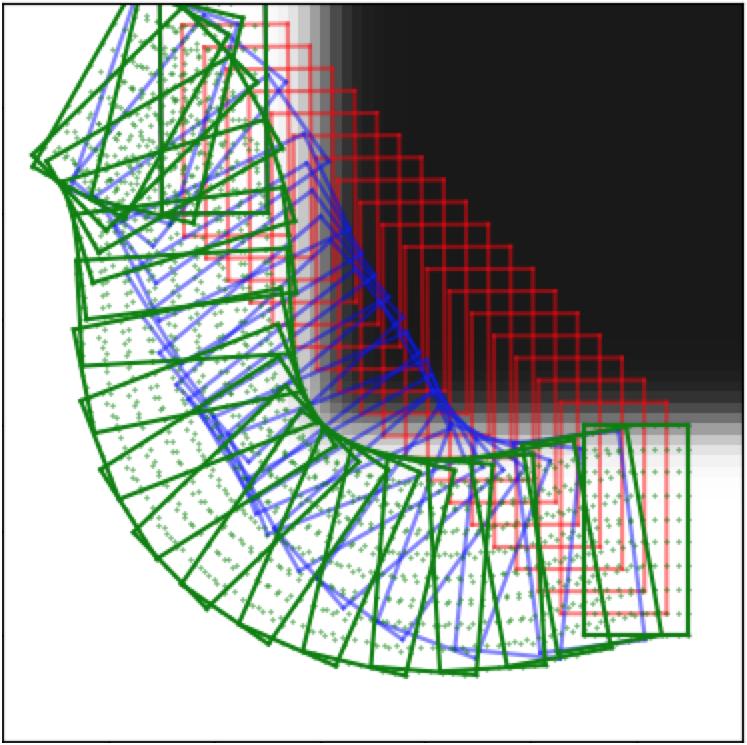}%
    \label{fig:2d_nerf_collision_avoidance}
}%
% \hfill
\subfigure[][]{%
    \includegraphics[clip, trim=2cm 0cm 3cm 0cm, width=0.45\linewidth]{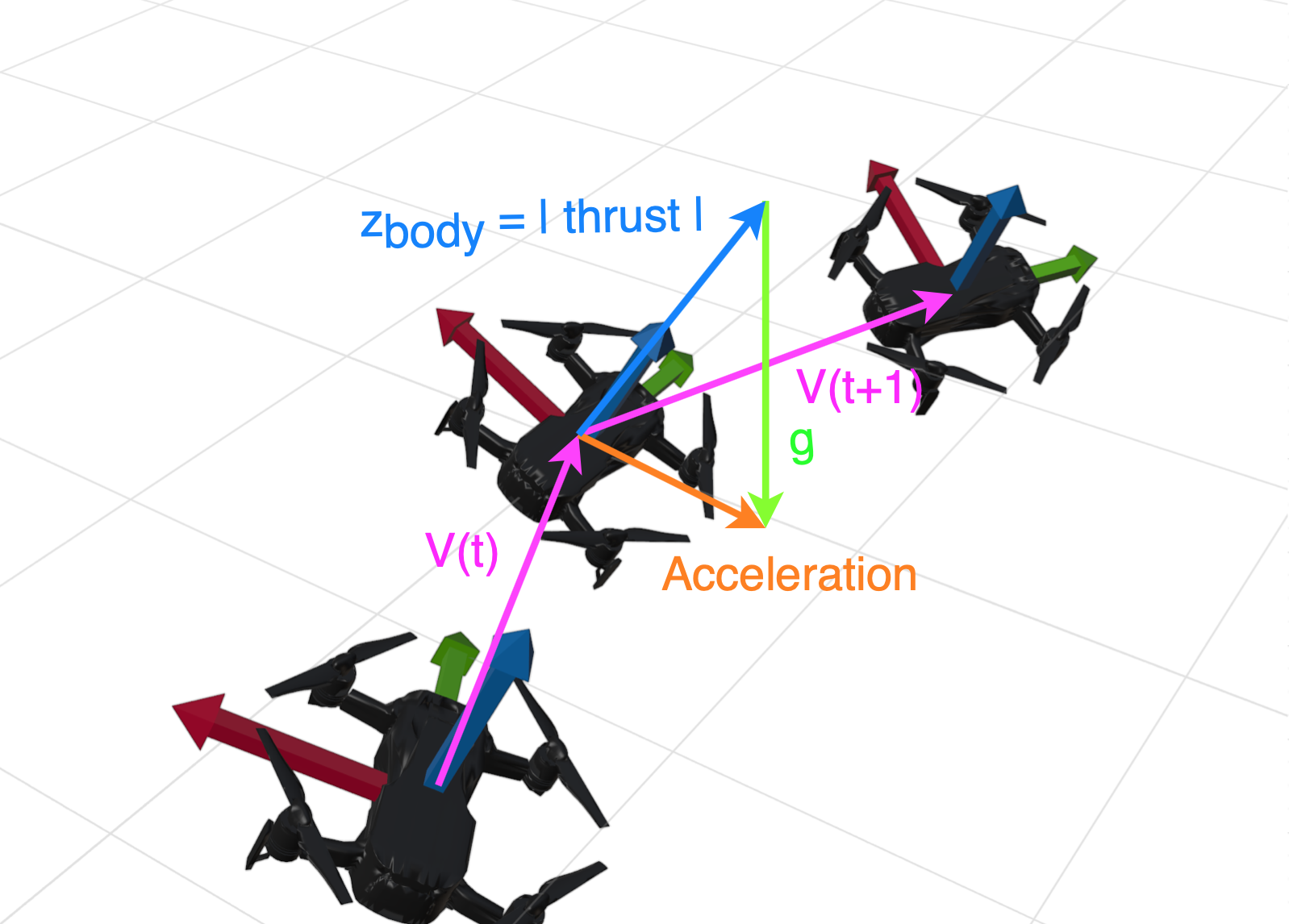}%trim=left bottom right top % \raisebox{3mm}{} around \includegraphics to raise an image if needed
    \label{fig:diff_flatness}
}%
\caption{\subref{fig:2d_nerf_collision_avoidance} Overhead view showing the planned trajectory as the optimization progresses on a toy example. The initial trajectory (red) goes straight through the high density regions of the \ac{nerf} (black area). The blue is an intermediate trajectory which clips the corner. The final trajectory (green) avoids the obstacle. %
\subref{fig:diff_flatness} Visualization of how the force balance on the quadrotor leads to a differential flatness formulation}

\end{figure}

The trajectory optimizer seeks a set of flat output waypoints $W = \left\{\sigma_0, \ldots, \sigma_h\right\}$ that minimizes multi-objective cost given by
\begin{equation}
J(W) = \sum_{\tau=0}^h  \Biggl[\ \ \overbrace{\sum_{b_i \in \mathcal{B}}\rho(R_\tau b_i + \hat{\sigma}_\tau) s(b_i)  }^{\text{collision penalty}} + \overbrace{\vphantom{\sum_{b_i \in \mathcal{B}}} u_{\tau}^T \Gamma u_{\tau} }^{\text{control penalty}}\Biggr]
\label{eqn:cost_fun_traj}
\end{equation}
% \begin{equation}
% \begin{split}
% J(W) = \sum_{\tau=0}^h &  \underbrace{ \left[k_i\left(\sum_{b_i \in \mathcal{B}} \rho(b_i)\right) \| v_{\tau} dt \|_2 \right. }_{\text{collision penalty}}  \\ & + \underbrace{\left. \vphantom{\left(\sum_{b_i \in \mathcal{B}}b_i N(\hat{\sigma}_{\tau})\right)}\gamma u_{\tau} \right]}_{\text{control penalty}}
% \end{split}
% % Note: added a `\vphantom` to the second term to align the underbraces vertically -- PC.
% \label{eqn:cost_fun_traj}
% \end{equation}

where $\hat{\sigma}_\tau$ is the position component of a differentially flat state $\sigma_\tau$, $R_\tau$ is the rotation matrix from the robot body frame to the world frame, $s(\cdot)$ is a function that returns the distance traveled by the point in the robot's point cloud, $\Gamma$ is the diagonal matrix of of weights penalizing control effort. The first  objective seeks to minimize the probability of collision, as defined in \eqref{eqn:collision_prob}, and the second seeks to minimize control effort. Note that $R_\tau$, $s(\cdot)$, and $u$ are derived from the surrounding waypoints using the robot's dynamics, and therefore are functions of the decision variables $\left\{\sigma_1, \ldots \sigma_h\right\}$. % $k_i$ is an optional weighting term that is a function of optimization iteration (see \eqref{eqn:collision_sigmoid})

%This cost function can be optimized using the Adam optimizer via tools provided by PyTorch or other machine learning libraries. In order to improve convergence the collision penalty term in the loss function can optionally be weighted by a scalar that is a function of the optimization index and trajectory step so that the planner initially ignores collisions towards the end of the trajectory and instead prioritize avoiding earlier collisions. This can help prevent the planner from getting stuck if the trajectory initialization is very poor. %As the optimization continues all collisions should be considered to avoid infeasible routes. 

% modified by multiplication via a sigmoid function of the optimization iteration index $i$, represented in Eqn.\ref{eqn:cost_fun_traj} as 
% % $k_i$.
% \begin{equation}
% k_i = 
% \begin{cases}
% \qquad\ \ 1, & i \geq \alpha_1 \\
% \frac{1}{1+e^{-\alpha_2(\tau/h - i/\alpha_1)}}, & i < \alpha_1
% \end{cases}
% \label{eqn:collision_sigmoid}
% \end{equation}
% where $\tau$ is the waypoint index, $h$ is the total number of waypoints, $i$ is the optimization step, $\alpha_1$ defines the training epoch after which this term is always 1, and $\alpha_2$ controls how sharp of an effect this term has. Effectively, these hyper parameters allows the planner to initially ignore collisions that would occur towards the end of the trajectory, and prioritize avoiding earlier collisions. As the optimization continues all collisions will be considered to avoid infeasible routes. 

\subsection{Initialization}

Our method is initialized by calculating a series of preliminary waypoint poses between the current pose and goal pose via a heuristic, such as a straight line or A$^*$ on a coarse grid overlaid on the scene. We optimize these initial guesses via gradient decent to balance multiple objectives such as avoiding collisions, and minimizing control effort.  Fig.~\ref{fig:2d_nerf_collision_avoidance} shows a trajectory moving towards areas of low \ac{nerf} density as it its optimized from an initial straight line. %In Sec.~\ref{Sec:RecedingHorizon} we show how this planner can be combined with a state estimation filter in a receding horizon manner to account for deviations from the planned trajectory. 

\section{Vision-Only Pose Filtering in a \ac{nerf}}
\label{Sec:StateEstimation}
After executing an action from the planned trajectory, the robot must close the loop and estimate its pose using its on-board sensors (e.g. a monocular camera). In this section we address the problem of how a robot can update its pose belief given a measurement and its most recent control action.

Our method is most closely related to \cite{yen2020inerf} where an initial pose estimate is optimized by minimizing the photometric loss between the pixels in the image and the predicted pixels via the projected \ac{nerf} scene. However, this method is a single-shot estimator and is highly dependent on the initialization. We formulate a state estimation filter that adds a process loss to the same photometric loss. This additional loss term provides benefits beyond the prior work by estimating a pose and its derivatives. Additionally, the state estimation should be more robust when the robot travels through regions of low photometric gradient information, relying more on the dynamics model. Lastly, the filter produces a state covariance which can be useful for other robotics algorithms running in parallel, such as collision avoidance with dynamic agents \cite{shah2019collisionavoid}.

\subsection{Optimization formulation}

At each timestep, the estimator is provided a new image $I_t$, the previous action taken $u_t$, a state prior $\mu_{t-1}$, and covariance $\Sigma_{t-1}$ of the previous time step. We propagate both the uncertainty and estimate as follows:
\begin{align}
\mu_{t|t-1} &= f(\mu_{t-1}, u_t) \\
A_{t-1} &= \frac{\partial f(x, u_t)}{\partial x}\Bigr|_{x=\mu_{t-1}} \\
\Sigma_{t|t-1} &= A_{t-1} \Sigma_{t-1} A_{t-1}^T + Q_{t-1}
\label{eqn:dynamcis_pose_est}
\end{align}
% \begin{align}
% \tilde{\mu}_t &= f(\mu_{t-1}, u) \\
% A_{t-1} &= \frac{\partial f(x)}{\partial x}\Bigr|_{x=\mu_{t-1}} \\
% \tilde{\Sigma}_t &= A_{t-1} \Sigma_{t-1} A_{t-1}^T + Q_{t-1}
% \label{eqn:dynamcis_pose_est}
% \end{align}
where the dynamics are modeled as $x_t = f(x_{t-1}, u_t)$ with process noise covariance $Q_t$.

\edit{As in \cite{yen2020inerf}, a subset of pixels $\mathcal{I}$ are selected for evaluation using existing image feature detectors (e.g. ORB) to identify points of interest and bias the sampling around these areas of higher gradient information.} The pose of the robot $T_t$ can be constructed from the position and rotation elements of $\mu_t$. With this information, the cost function to be minimized is 
\begin{equation}
J(\mu_t) = \overbrace{\|C_{\mathcal{I}}(T_t) - I_t(\mathcal{I}) \|^2_{{S^{-1}_t}}}^{\text{photometric loss}} + \overbrace{\|\mu_{t|t-1} - \mu_{t} \|^2_{\Sigma_{t|t-1}^{-1}}}^{\text{process loss}}
\label{eqn:cost_fun_pose}
\end{equation}
where $S_t$ is the measurement noise covariance and the notation $\|x\|^2_M = x^TMx$ is the weighted $\ell_2$ norm. Minimizing this equation gives the updated mean $\mu_t$. \edit{Outlier rejection is performed on the per-pixel loss to reduce variance.

Finally, we leverage the known relationship between the Hessian of a Gaussian loss function and the covariance \cite{yuen2010} to yield the posterior covariance,}
\begin{equation}
\Sigma_t = \left(\frac{\partial^2 J(x)}{\partial x^2}\Bigr|_{x=\mu_t}\right)^{-1}.
% \Sigma_t = \left(\frac{\partial^2 J(\mu_t)}{\partial x^2}\right)^{-1}.
\label{eqn:pose_covar}
\end{equation}

\subsection{Performance enhancing optimization details}

%{To improve estimation performance, we optimize $J(\cdot)$ over a vector of deviations from the predicted state $\mu_{t|t-1}$. $9$ of the $18$ elements of the state vector are the rotation matrix elements representing the orientation of the robot. Since we want to optimize directly on $\text{SO}(3)$, the rotation delta is a vector in $\mathbb{R}^3$ denoted $\delta$, which lies in the tangent space of $\text{SO}(3)$ at $R_{t-1}$. After the optimization, the exponential map denoted $\exp(\cdot)$ \cite{sola2018micro} can be used to update the rotation matrix,

%\begin{equation}
%R_t = R_{t - 1} \exp(\delta).
%\label{eqn:exp_map_update}
%\end{equation}

\begin{figure}[t]
\centering
\includegraphics[clip, trim=2.5cm 6.25cm 5cm 8.25cm, width=0.90\linewidth]{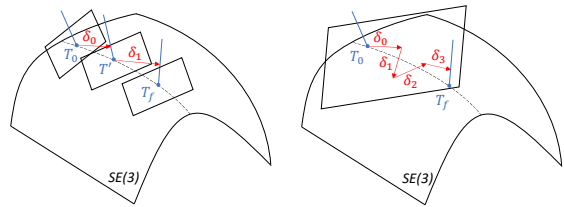} %trim=left bottom right top
\caption{\edit{Recursive $SE(3)$ optimization (left) vs. Optimization in tangent plane (right)}}
\label{fig:tangent_planes}
\end{figure}

\edit{ % begin lie edit
The approach in \cite{yen2020inerf} optimizes for the state by taking gradient steps with respect to a reference pose and projecting onto the $\text{SE}(3)$ manifold after the optimization to recover the state estimate. Instead, we project back onto the manifold after every gradient step. This is illustrated in Fig.~\ref{fig:tangent_planes}. These two methods are mathematically distinct, as the multiplication of skew-symmetric matrices in the exponential map do not commute. Explicitly,
\begin{equation}
\exp(\delta_1) \exp(\delta_2)\cdots  T \neq \exp(\delta_1 +\delta_2 + \cdots ) T,
\label{eqn:exp_map}
\end{equation}
where T is the reference homogeneous transformation matrix, and $\exp(\cdot)$ is the exponential map between the tangent space and manifold. Qualitatively we observe that the recursive SE(3) gradient descent converges quicker and more smoothly than the method in \cite{yen2020inerf} because of the noisy photometric loss landscape over the $\text{SE}(3)$ manifold. Please see \cite{sola2018micro} for further Lie theory details. Optimization on the manifold is implemented using the LieTorch library \cite{teed2021lietorch}.
} % end lie edit

\section{Online Replanning for Vision-Only Navigation}
\label{Sec:RecedingHorizon}

We combine the trajectory planner from Sec.~\ref{Sec:TrajectoryPlanning} and the state estimator from Sec.~\ref{Sec:StateEstimation} in an online replanning formulation. The robot begins with an initial prior of its state belief, a final desired state, as well as the trained \ac{nerf}. 

\begin{figure}[]
    \centering
    \includegraphics[clip, trim=1.75cm 2cm 2cm 2cm, width=0.9\linewidth]{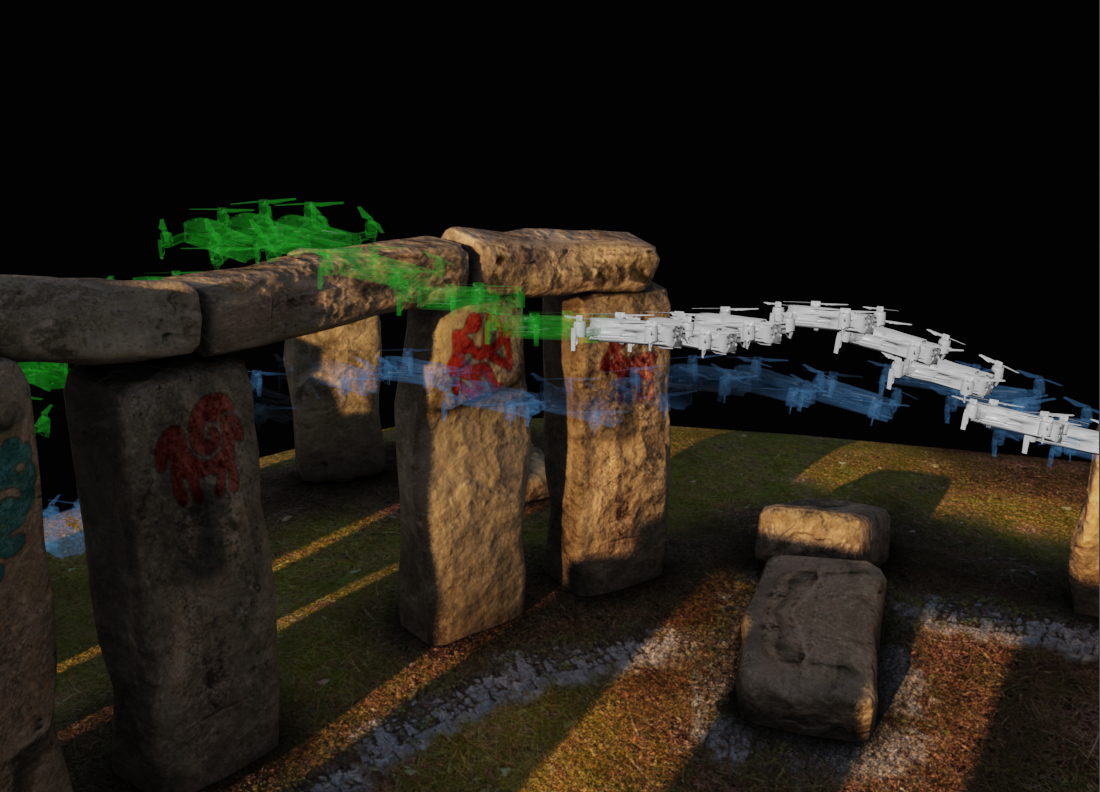}%trim=left bottom right top
    \caption{Results for the proposed trajectory optimizer navigating through Stonehenge. The blue trajectory is the initial plan returned by our optimizer. We then roll out the dynamics noisily for a number of timesteps (white) and re-optimize the trajectory (green). The planner responds to a vertical disturbance by opting to fly above the arch, rather than below as initially planned. Changing homotopy classes around obstacles is quite difficult for existing differential flatness-based planners }
    \label{fig:stonehenge_traj}
\end{figure}

The robot first plans a trajectory as described in Sec.~\ref{Sec:TrajectoryPlanning}. The robot then executes the first action (in this case \edit{inside} a simulator), and the state filter takes in a new image and updates its belief. The mean of this posterior is used in the trajectory planner as a new starting state and along with the rest of the previous waypoints at a hot start, re-optimizes the trajectory taking into account any disturbances. This continues until the robot has reached the goal state. \edit{ This process is described in Alg.~\ref{algo:main}. This allows the robot to create new updated plans that take into account disturbances. Figure \ref{fig:stonehenge_traj} show an example of a trajectory being reoptimised given new information.}

\begin{algorithm}
	\caption{Receding Horizon Planner}
	\label{algo:main}
	\begin{algorithmic}[1]
		\State Inputs: ($\mu_0$, $\Sigma_0$) initial state prior, $x_{goal}$ desired final state, $N$ trained \ac{nerf} model of env.
        \State $W \leftarrow$ \texttt{A}$^*(\mu_0, x_{goal})$ 
		
		\While {not at $x_{goal}$}
    		\State $W \leftarrow \texttt{trajOpt}(W)$ [Sec.~\ref{Sec:TrajectoryPlanning}]
    		\State $\mathbf{x}$, $\mathbf{u} \leftarrow \texttt{getStatesActions}(W)$ %[Eqn.~\ref{eqn:vel_and_accel_traj}-\ref{eqn:contol_input}]
    	    \State $I \leftarrow \texttt{\edit{getCameraImage}}()$
    		\State $\mu_t, \Sigma_t  \leftarrow \texttt{poseFilter}(I, \mu_{t-1}, \Sigma_{t-1}, \mathbf{u}[0])$  [Sec.\ref{Sec:StateEstimation}]
    		\State $W \leftarrow [\mu_t, W\texttt{[2:end]}]$
		\EndWhile

	\end{algorithmic} 
\end{algorithm}

\section{Experiments}
\label{Sec:Experiments}

We demonstrate the performance of our method using a variety of high-fidelity simulated mesh environments (scene meshes by Sketchfab users Ahmad Azizi, artfletch, \& ruslans3d). Since our method assumes a trained NeRF model, we first render a sequence of images from the mesh. These images are used to train a NeRF model using an off-the-shelf PyTorch implementation \cite{lin2020nerfpytorch}. Rendering images from a mesh with \cite{szot2021habitat} (rather then images taken with a camera in the real world) provides a ground truth reference for the scene geometry with which to evaluate our method. \edit{ The robot's sensor images are similarly rendered from the ground truth environment, but the trajectories are different so any query of the NeRF model differs from the training data.}

\edit{The experiment section is divided into 3 parts: We first evaluate the performance of our trajectory planner on its own, then the state estimator on it own, before demonstrating the complete online replanning pipeline.}

% Note: whitespace can affect the positioning of the figures (if you have line breaks it might make the figures vertical instead of horizontal 
% \begin{figure*}[t]
% \centering
% \subfigure[][]{%
% \includegraphics[width=0.45\linewidth]{example-image-a}
% \label{fig:sub_fig_a}%
% }
% \hspace{2pt}%
% \subfigure[][]{%
% \includegraphics[width=0.45\linewidth]{example-image-b}
% \label{fig:sub_fig_b}%
% }
% \caption{This figure spans two columns}
% \label{fig:two_col_fig}
% \end{figure*}

% \subsection{Trajectory Planning}
We first study the performance of our trajectory optimizer alone on a number of benchmark scenes. We demonstrate that our trajectory optimizer can generate aggressive, dynamically feasible trajectories for a quadrotor and an omni-directional robot which avoid collision. 

% \st{To evaluate how well the robot avoids collisions with the environment, we use compare the same collision penalty metric used in our loss function in Eqn.~\ref{eqn:cost_fun_traj} with the intersection volume between the robot mesh and the true underlying mesh along the same trajectory. }

\subsection{ \edit{Planner - Ground truth comparison} }

\begin{figure}[ht]
\centering
% \hspace*{\fill}%
\subfigure[][]{%
\includegraphics[width=0.35\textwidth]{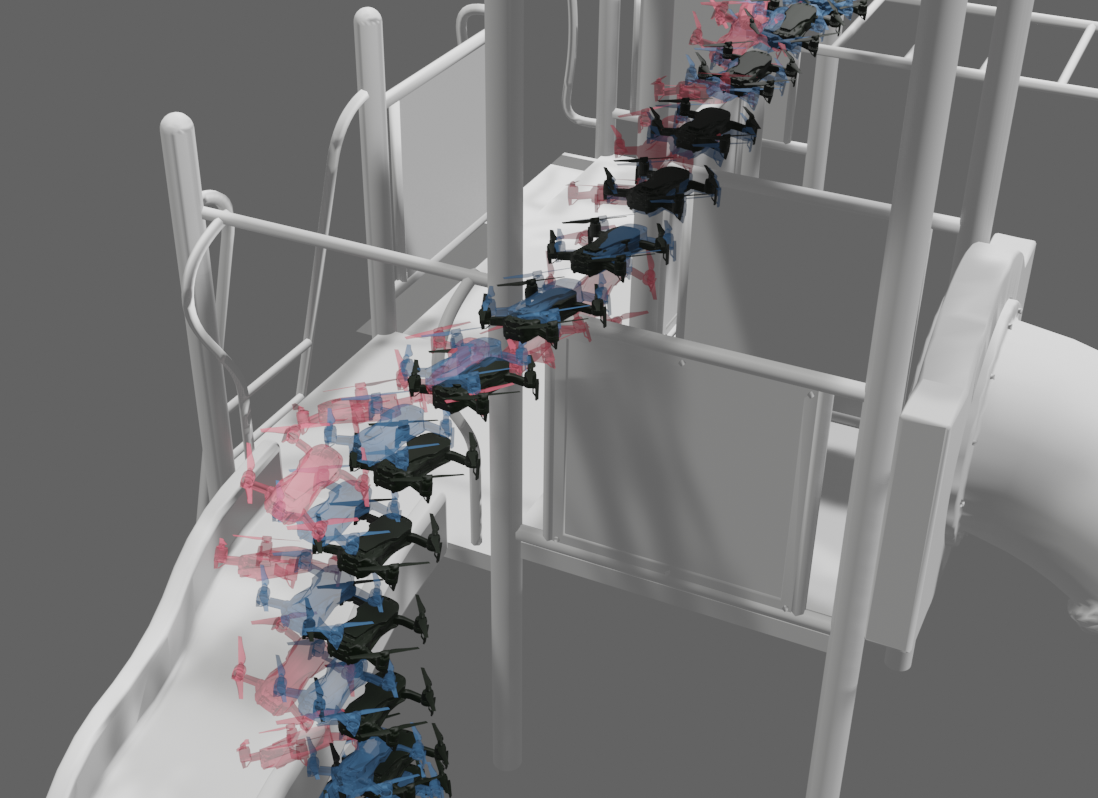}%
\label{subfig:traj_playground_view}
}%
\vfill
\subfigure[][]{%
\includegraphics[width=0.4\textwidth]{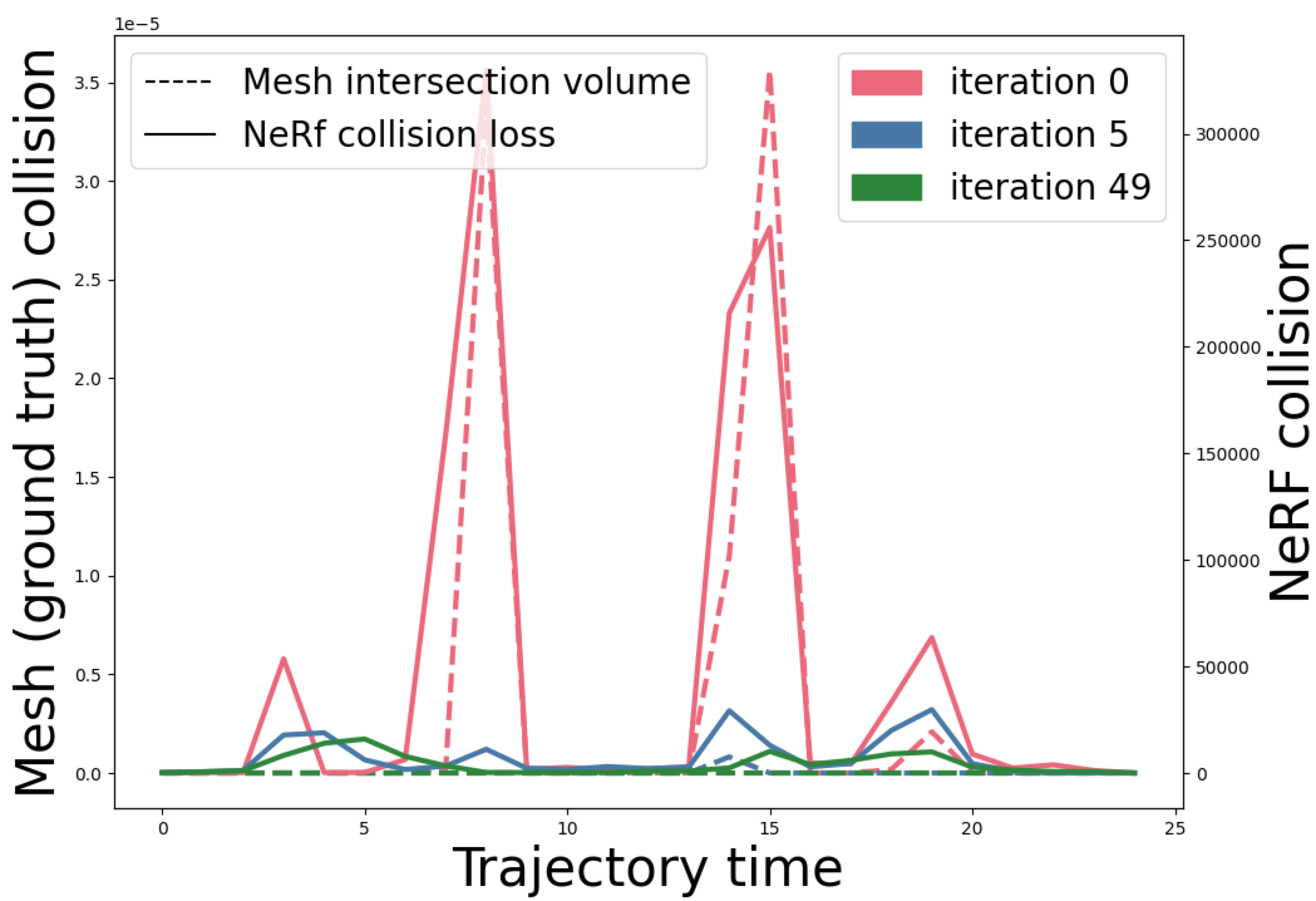}%
\label{subfig:traj_playground_meshplot}
}
% \hspace*{\fill}%

\caption{Results of our proposed trajectory optimizer planning a path through a playground. \textbf{(a)} Visualization of the optimized trajectory generated by our planner. The initialization provided is shown in red and a partially-optimized trajectory in blue. We see the optimizer converge to a trajectory that both avoids collision and is smooth by observation. \textbf{(b)}: Plot of the \ac{nerf} collision loss (solid lines), and the intersecting volume of the ground-truth meshes (dashed lines). Lower is better. We see the \ac{nerf} collision loss is clearly correlated with the intersection volume, showing that minimizing our proposed objective \eqref{eqn:cost_fun_traj} indeed minimizes collision. Note that by iteration 49 optimizer converges to a trajectory that has zero intersections with the ground truth meshes.} 
\label{fig:quadrotor_trajectory_playground}
\end{figure}

\edit{ In order to use a \ac{nerf} to reason about collisions, we need to show that the learned optical density is a good proxy for real world collisions. We compare the \ac{nerf} predicted collisions during various stages of trajectory optimization with the ground truth mesh intersecting volume, during planning of a quadrotor path through a playground environment. The trajectories are shown in Fig.~\ref{subfig:traj_playground_view}. } Fig.~\ref{subfig:traj_playground_meshplot} shows the relationship between the collision loss term from \eqref{eqn:cost_fun_traj} and overlap between the robot volume and the ground truth mesh \edit{over time}. \edit{In addition to planner finding a smooth, collision-free (i.e., zero mesh intersection volume) trajectory, we can see that throughout training the \ac{nerf} density and mesh overlap are closely matched.}

% The solid robot figures represent the optimized waypoint poses, while the translucent images show a snapshot of the partially optimized trajectory. 

% The solid robot figures represent the optimized waypoint poses, while the translucent images show a snapshot of the partially optimized trajectory. 

% \begin{figure}[h]
%     \centering
%     \includegraphics[width=0.95\linewidth]{figures/gt_compare/planner_random_points_anotated.png}%trim=left bottom right top
%     \caption{The start and endpoint of a range of trajectories, matched by color. The one trajectory that has a collision is marked with a cross.}
%     \label{fig:stonehenge_random_traj}
% \end{figure}

\subsection{\edit{Planner - Comparison to prior work }}

\begin{figure}[t!]
\centering
% \hspace*{\fill}%
\subfigure[][]{%
\includegraphics[width=0.35\textwidth]{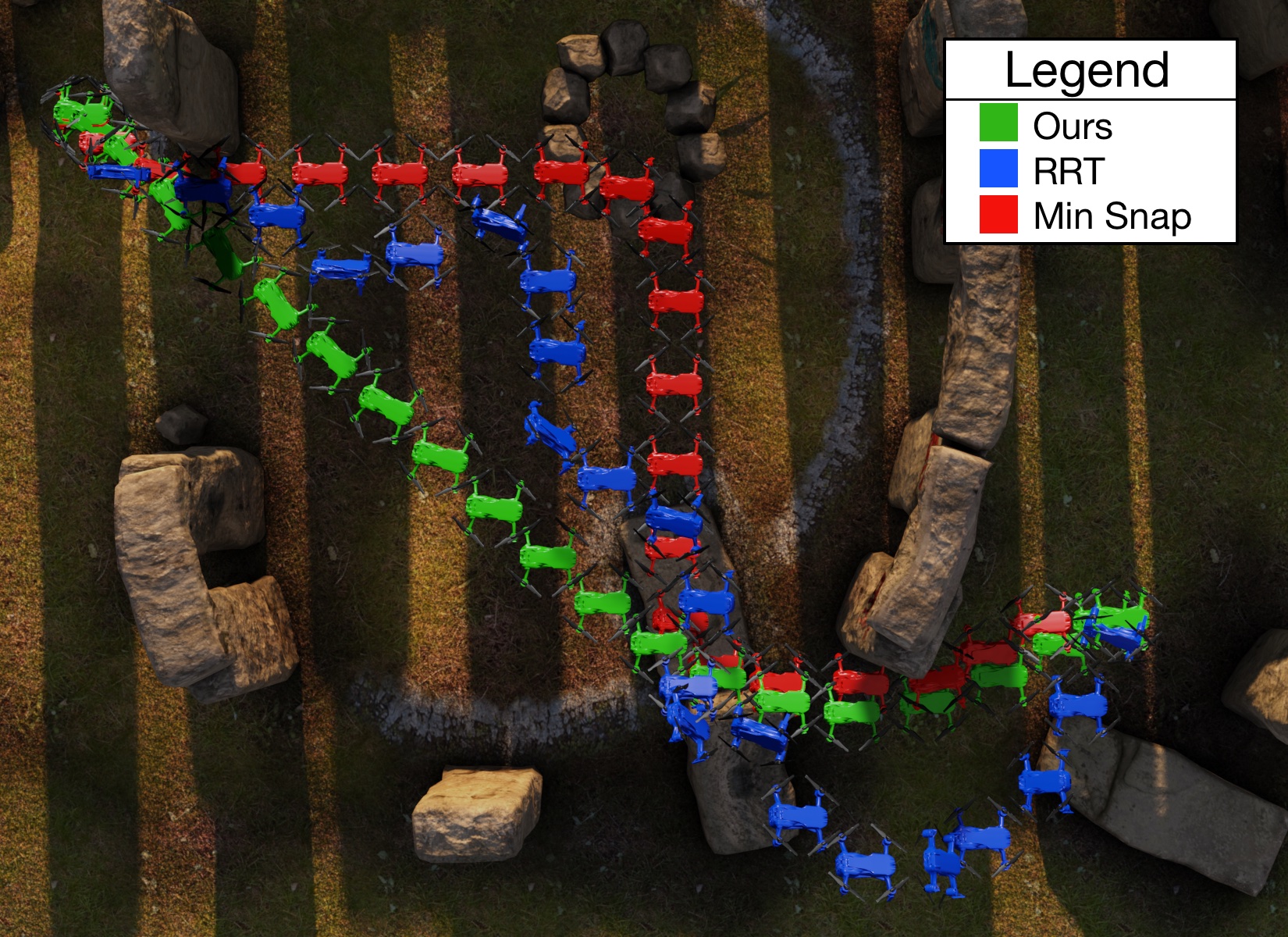}%
\label{subfig:traj_comp_view}
}\\%

\subfigure[][]{%
\includegraphics[width=0.35\textwidth]{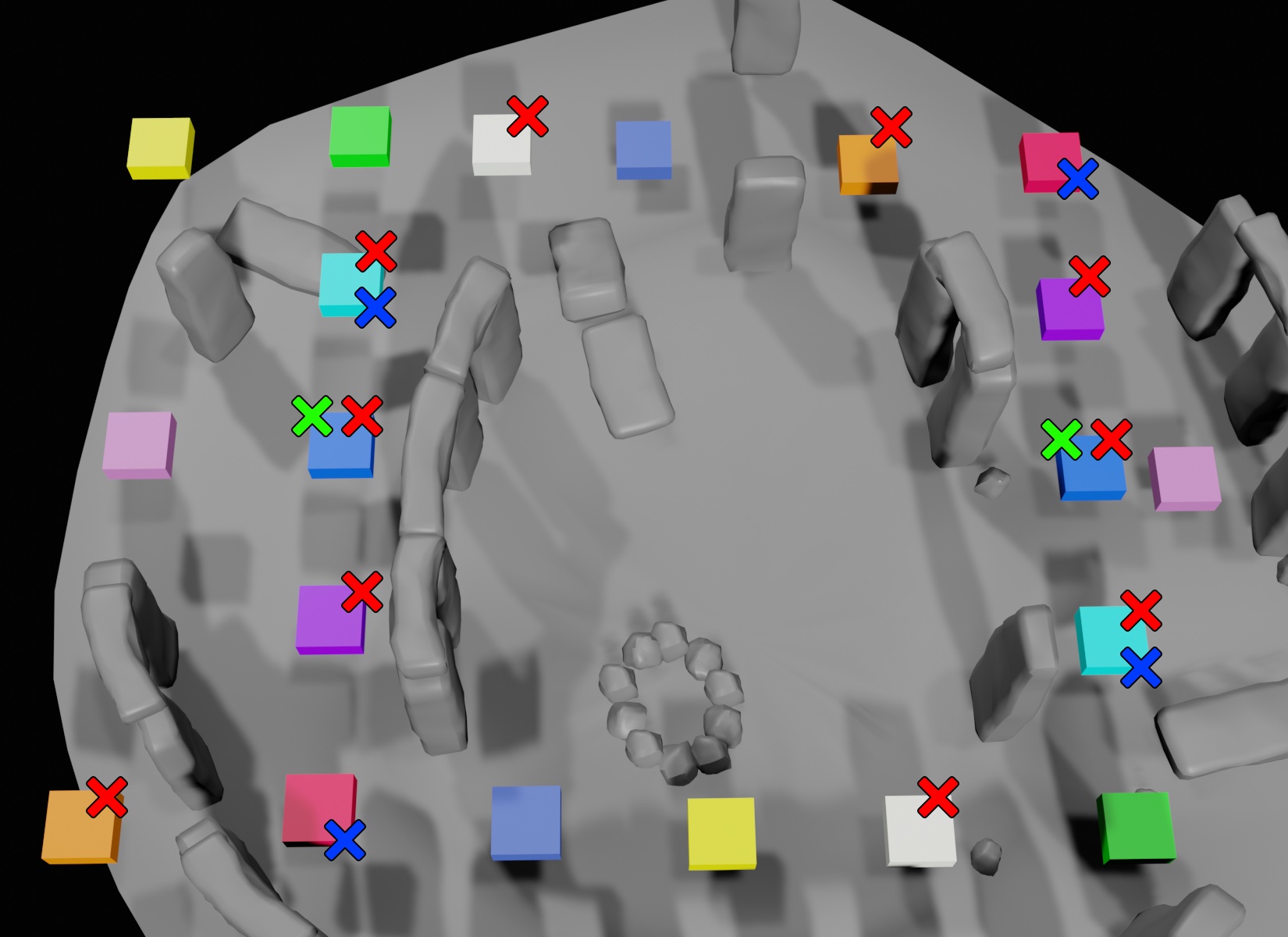}%
\label{subfig:random_points}
}\\

\subfigure[][]{%
\includegraphics[width=0.35\textwidth]{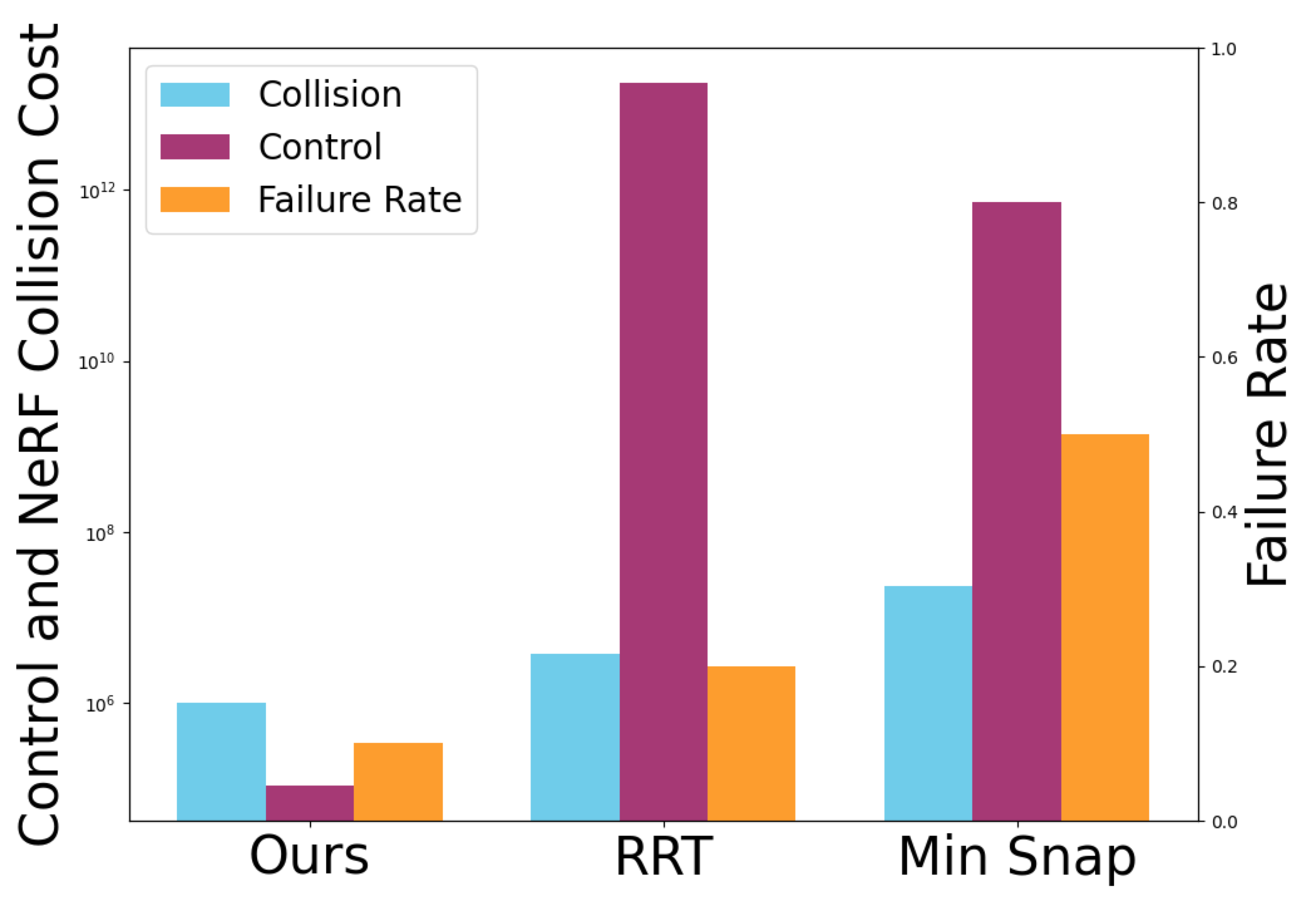}%
\label{subfig:traj_comp_graph}
% \hspace*{\fill}%
}
\caption{\edit{A comparison between our planner and the minimum-snap and RRT baselines.
\textbf{(a)} Example trajectories the planners take though the Stonehenge environment. Our planner can move the waypoints to result in a smooth trajectory compared to minimum-snap, which exactly follows the A* initialization. Further, while our planner's trajectory does not collide with the ground truth mesh the minimum-snap trajectory clips the column on the right. While RRT generates a collision-free trajectory, its erratic shape leads to a high control effort. \textbf{(b)} Color-matched start and endpoints of the trajectories along with an indication if they were successful for a given planner (crosses use the same coloring as in \textbf{(a)}). \textbf{(c)} Each planner's mean NeRF collision metric and control effort per time step, averaged over the 20 initializations. We can see that our method yields trajectories with low control effort, low NeRF collision cost, as well as a low failure rate.
}}

\label{fig:traj_comp}
\end{figure}

\edit{Since this method is, to our knowledge, the first method to operate on NeRFs, direct comparisons are difficult. We compare to two widely used techniques that we have adapted to work on a \ac{nerf} environment representation: minimum-snap trajectory planning and Rapidly-exploring Random Trees.

Minimum-snap trajectory planning \cite{mellinger2012}, similarly to our method, uses differential flatness to compute trajectories that pass through a series of waypoints. However, this method typically uses hand-placed waypoints, whereas our method is capable of optimizing the locations of those waypoints based on the \ac{nerf}. In this comparison, we generate the waypoints for the minimum-snap planner using the same A* algorithm our method uses.

Rapidly-exploring Random Trees (RRT) is a sampling based method that generates a space-filling tree used to find a collision free trajectory. Since it requires a binary collision metric, we first convert the \ac{nerf} into a mesh using marching cubes, as in \cite{mildenhallNeRFRepresentingScenes2020a}. When generating the RRT, we use a spherical collision model, as we cannot know the robot's orientation at the planning stage, since it selects only positions. Finally, in order to extract the control actions required to follow the RRT trajectory, we use a a differential flatness-based controller \cite{mellinger2012}. 

To evaluate performance we run all 3 methods on 10 trajectories with a range of obstacles, speed and complexities inside the Stonehenge environment. Fig.~\ref{subfig:traj_comp_graph} shows the mean costs associated with each planner, along with the failure rate (defined as an collision with the ground truth mesh) in the trajectories. Additionally Fig.~\ref{subfig:traj_comp_view} shows the 3 planners' qualitative performance. }

\subsection{\edit{Planner - Omnidirectional robot in tight space}}

\begin{figure}[t!]
    \centering
    \includegraphics[clip, trim=0cm 0cm 0cm 0cm, width=0.8\linewidth]{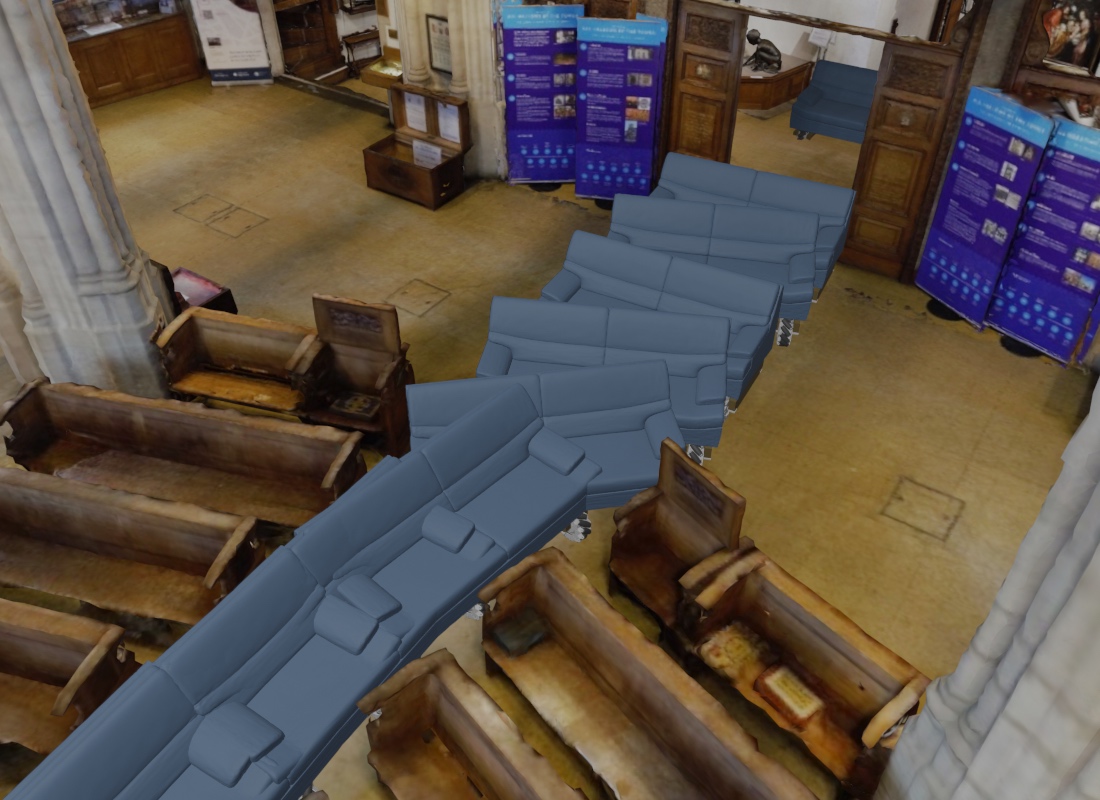}%trim=left bottom right top
    \caption{A planned trajectory for a couch-shaped mobility robot through a narrow gap. The proposed \ac{nerf}-based collision penalty results in a trajectory which turns the robot to fit through the gap and avoid collision.
    }
    \label{fig:piano_mover_traj}
\end{figure}

Our method is not limited to  quadrotors, but can handle any robot with differently flat dynamics. Figure \ref{fig:piano_mover_traj} presents an omnidirectional, couch-shaped mobility robot navigating through a narrow space. This scenario presents a difficult kinematic planning problem, commonly called the ``piano movers' problem'' \cite{schwartz1983}, which requires the robot to turn to fit its body through the narrow gap. The trajectory optimizer, using the proposed \ac{nerf}-based collision penalty, is able to generate the desired behavior, which turns the robot to fit through the gap.

\subsection{\edit{Estimator - Comparison to prior work}}

\begin{figure}[]
    \centering
    \includegraphics[clip, trim=6cm 2cm 0cm 0.5cm, width=1.1\linewidth]{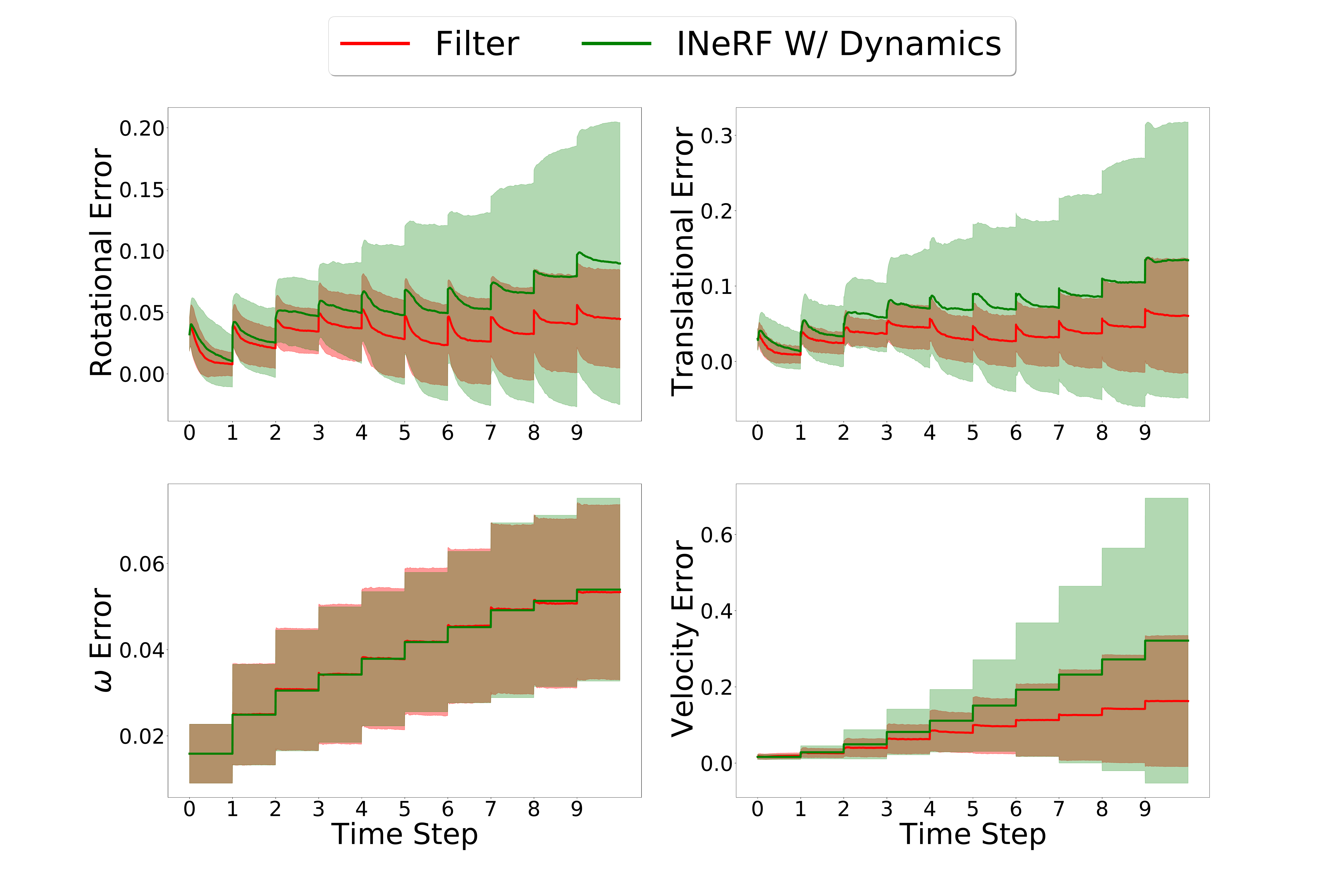}%trim=left bottom right top
    \caption{\edit{Error comparison averaged over 100 trials. Rotational errors are the $\ell_2$ norm of the angles in axis-angle representation required to rotate the estimated pose to the ground truth. Translational and rate errors are the $\ell_2$ norm of the estimated and ground truth difference. Bounds indicate one standard deviation above and below the mean error. Regions between time steps are the gradient steps of the optimization, while spikes at the beginning of time steps indicate the forward propagation of the simulation, a new image observation, and a new optimization. Our method (red) outperforms the dynamically initialized iNeRF (green) in rotational, translational, and velocity estimates while sporting lower variance.}}
    \label{fig:state_estimator_comparison}
\end{figure}

We evaluate two methods of estimating the robot state in the \ac{nerf} Stonehenge environment. We anticipate that an iNeRF \cite{yen2020inerf} estimator initialized without a dynamics prior will very quickly diverge. Therefore, we propose a dynamically-informed iNeRF estimator as a baseline. %\footnote{https://github.com/salykovaa/inerf/}. 
The estimate is propagated through the dynamics model to provide an initialization to the estimator at the next time step. Only the photometric loss is optimized.  This estimator is identical to a recursive Bayesian filter with infinite state covariance, hence the process loss is set to 0. The second method is our full filter proposed in Sec.~\ref{Sec:StateEstimation}. We evaluate these methods on a identical set of actions and initial state. For our filter, we assume $\Sigma_0 = 0.1 I$, $Q_t = 0.1 I$, $S_t = I$, and use $\mathcal{I} = 256$ pixels. Zero-mean Gaussian white noise is added to the true dynamics with standard deviation $2 cm$ for the translation and $0.02 rad$ to the pose angles, while the standard deviation for their rates are half those values. For comparison, the scene area is scaled to be approximately $4 m^2$ and the drone is $0.5 cm^3$ in volume.

\edit{A comparison on the two methods over 100 trials conditioned on the same initial state, set of actions, and noise characteristics is shown in Fig.~\ref{fig:state_estimator_comparison}. Our method outperforms the dynamically-informed iNeRF baseline on almost every metric and does not under-perform. We again bring attention to the fact that our filter provides a finite state covariance, which may be useful in determining low-fidelity regions of the NeRF environment.}
%\footnote{We used an unofficial implementation \cite{lin2021inerfcode} as no official one has been released.}

% How much noise is injected?
% - solid poses have already been executed and future plans at various time steps are faded.

\subsection{ \edit{Online Replanning} }
\begin{figure}[t]
\centering  
\hspace*{\fill}%
\subfigure[][]{%
\includegraphics[width=0.23\textwidth]{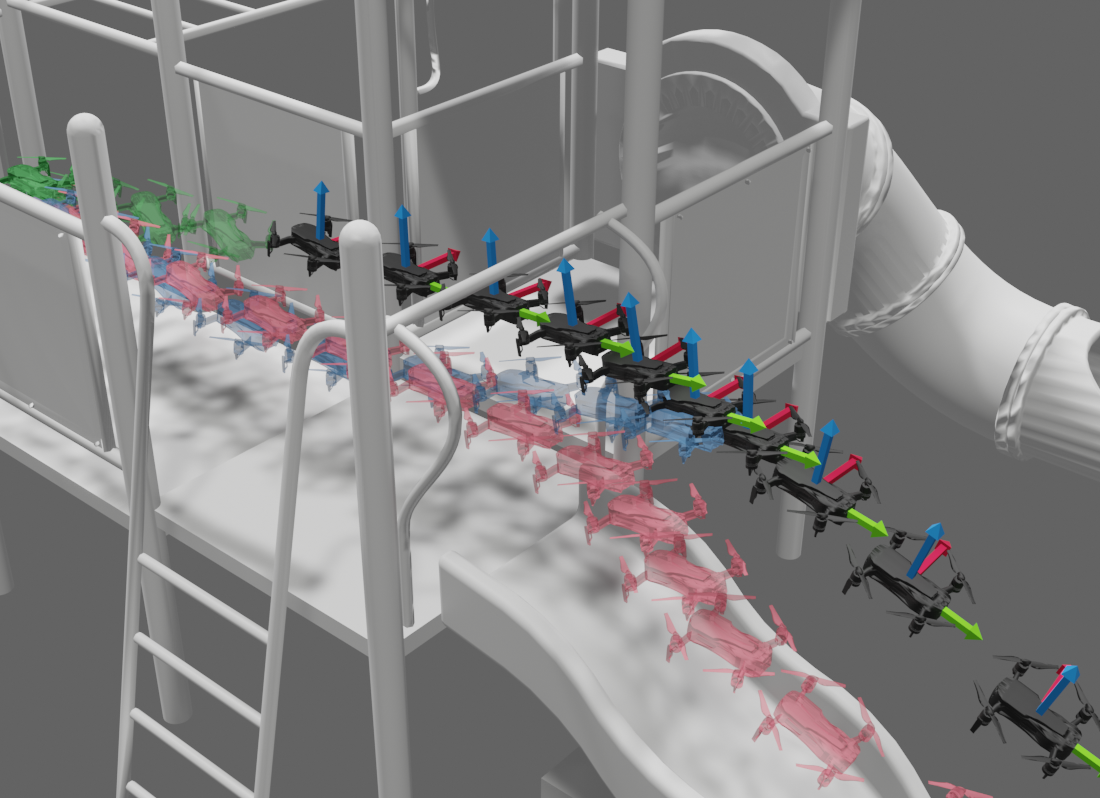}%
\label{fig:traj_playground_view_mpc}
}%
\hfill
\subfigure[][]{%
\includegraphics[width=0.23\textwidth]{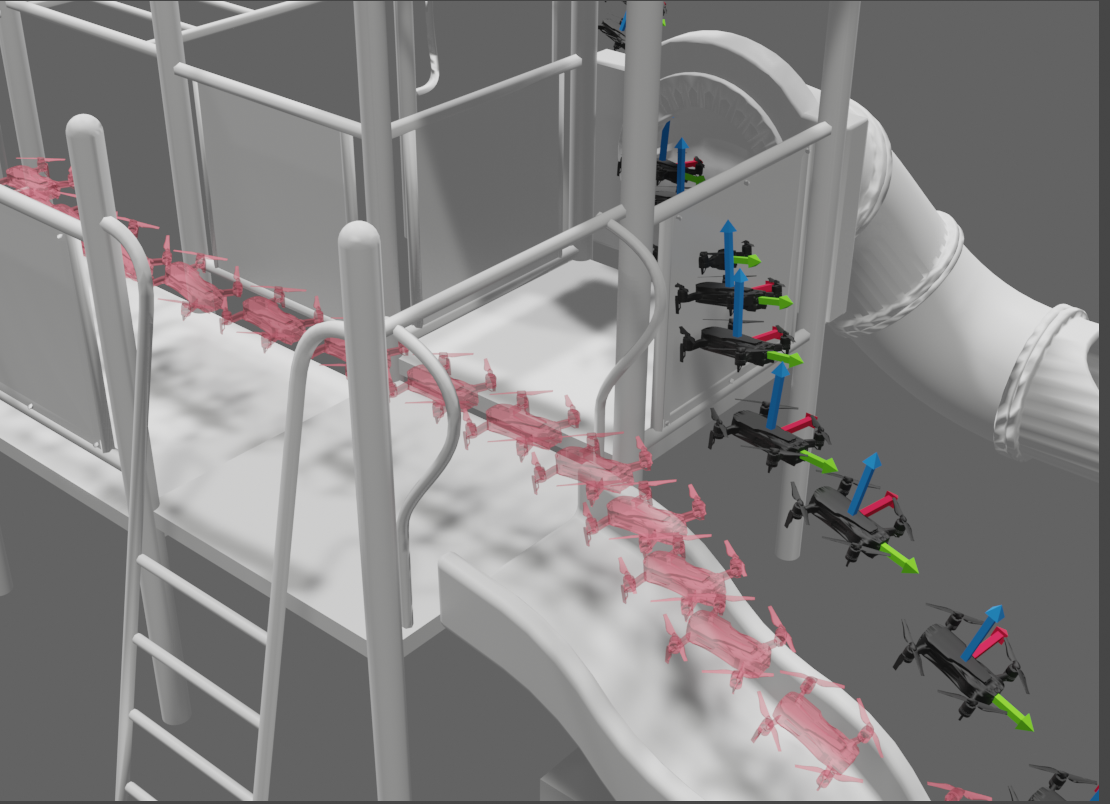}%
\label{fig:traj_playground_view_openloop}
}%
\hspace*{\fill}%

% \vskip\baselineskip

% \hspace*{\fill}%
% \subfigure[][]{%
% \includegraphics[width=0.4\textwidth]{figures/seperated_cost_fig_test.png}
% \label{fig:playground2_openloop}
% }%
% \hfill
% \subfigure[][]{%
%     \includegraphics[width=0.4\textwidth]{figures/tmp_placeholder.png}
%     \label{fig:openloop_cost}
% }%
% \hspace*{\fill}%

\caption{
\textbf{(a)} Quadrotor flight path execution with feedback. The originally planned trajectory is in red. However, when the state estimate deviates significantly from the planned trajectory, the robot re-plans and executes a collision-free path to the goal, as shown by the re-planned trajectories in blue and green.
\textbf{(b)} Quadrotor flight path execution without feedback (open-loop). An external disurbance causes the trajectory to deviate from the original plan (red) with catastrophic results.}

\end{figure}

We evaluate performance of the entire pipeline on planned trajectories in the playground and Stonehenge scenes. The ground truth dynamics are the finite difference drone equations in Sec.~\ref{Sec:StateEstimation} with the same additive noise as in our estimator experiment. Although the executed trajectories incur a higher cost than the initial plan, the planner is still able to generate collision-free trajectories (Fig.~\ref{fig:traj_playground_view_mpc}) and reach the goal, whereas an open-loop execution (Fig.~\ref{fig:traj_playground_view_openloop}) of the initial planned actions causes collisions and divergence.

\subsection{ \edit{Performance and Timing} }
% (Table of speeds)
% \begin{center}
% \begin{tabular}{|l|c|c|c| } 
%  \hline
% Scene & Playground & Stonehenge & Church  \\
%  \hline
%  Planner iteration & X ms & 26ms & X ms  \\ 
%   \hline
%  New plan          & X ms & 67s  & X ms \\ 
%   \hline
%  State estimation  & X ms & X ms & X ms \\ 
%   \hline
%  Online replan    & X ms & X ms & X ms \\ 
%  \hline
% \end{tabular}
% \end{center}

\edit{Experiments were run on a computer with an AMD Ryzen 9 5900X @ 3.7 GHz CPU and Nvidia RTX 3090 GPU. Both the trajectory planner and estimator computation time is dependent on number of iterations. Typically, an initial trajectory requires 20s over 2500 iterations to optimize. In the online replanning experiments ($\Delta t = 0.1s$), subsequent trajectory updates occur in 2s over 250 iterations. The state estimator typically runs for 4s over 300 gradient steps, 0.25s of which is the Hessian computation \eqref{eqn:pose_covar}. However, NeRFs are a fast-evolving technique and extensions have seen orders-of-magnitude improvements in performance \cite{garbin2021fastnerf, sun2021direct}, which we hope to leverage in the future.

Assuming those performance gains apply for our use case, we could expect to be able to run this type of algorithm in real time on a real robot, perhaps aided by off-board compute, in the near future.}

% Recently, \acp{nerf} have shown impressive improvements in performance. Even the new version of the \ac{nerf} pipeline we use claims X performance improvements, and new papers such as X claim to be able to train a \ac{nerf} in 15 minutes using new architectures. 

%%%%
% \begin{figure*}[t]
% \centering
% \hspace*{\fill}%
% \subfigure[][]{%
% \includegraphics[width=0.4\textwidth]{playground2_mpc_plot.png}%
% \label{fig:traj_playground_view_mpc}
% }%
% \hfill
% \subfigure[][]{%
% % \includegraphics[width=0.4\textwidth]{figures/playground2_mpc_cost.png}%
% \includegraphics[width=0.4\textwidth]{figures/seperated_cost_fig_test.png}%
% \label{fig:traj_playground_meshplot_mpc}
% }%
% \hspace*{\fill}%
% \caption{Quadrotor flight path execution with feedback. The originally planned trajectory is in red. However, when the state estimate deviates significantly from the planned trajectory, the robot re-plans and executes a collision-free path to the goal, as shown by the re-planned trajectories in blue and green.}
% \end{figure*}

% %%%%%
% \begin{figure*}[t]
% \centering
% \hspace*{\fill}%
% \subfigure[][]{%
% \includegraphics[width=0.4\textwidth]{figures/playground2_openloop.png}
% \label{fig:playground2_openloop}
% }%
% \hfill
% \subfigure[][]{%
%     \includegraphics[width=0.4\textwidth]{figures/openloop_cost.png}
%     \label{fig:openloop_cost}
% }%
% \hspace*{\fill}%
% \caption{Quadrotor flight path execution without feedback (open-loop). An external disurbance causes the trajectory to deviate from the original plan (red) with catastrophic results. }
% \end{figure*}

% %%%%
% \begin{figure*}[t]
% \centering
% \hspace*{\fill}%
% \subfigure[][]{%
% \includegraphics[width=0.4\textwidth]{figures/stonehenge_mpc_white2.png}
% \label{fig:stonehenge_mpc_white2}
% }%
% \hfill
% \subfigure[][]{%
%     \includegraphics[width=0.4\textwidth]{figures/stonehenge_mpc_cost.png}
%     \label{fig:stonehenge_mpc_cost}
% }%
% \hspace*{\fill}%
% \caption{Quadrotor flight path executed in a realistic outdoor environment with feedback. The high final cost (in green) is due to noise pushing the quadrotor downward, which results in a large thrust applied to reach the final state at the last moment.}
% \end{figure*}
\section{Conclusions}
\label{Sec:Conclusions}
In this work, we proposed a trajectory planner and pose filter that allow robots to harness the advantages of the \ac{nerf} representation for collision-free navigation. We presented a new trajectory optimization method based on discrete time differential flatness dynamics, and combined this with a new vision-based state filter to create a full online trajectory planning and replanning pipeline. 

Ongoing work seeks to further integrate perception and control in an active planning manner, both by encouraging the trajectories to point the camera in directions with greater gradient information as well as use the uncertainty metrics calculated by the state estimator to reduce collision risk. Another direction for future work includes harnessing improvements in the underlying \ac{nerf} representation to improve execution speed~\cite{garbin2021fastnerf}, since this is the limiting factor for the proposed method.

\edit{We also seek to extend this work to utilize multiple \acp{nerf} to represent scenes with movable objects, and explore how various robots such as manipulators could interact with such an environment.} Lastly, further work could look to improve the pixel sub-sampling heuristic employed by the state filter. Finally, we would like to implement the proposed method on quadrotors in real scenes to demonstrate the performance beyond simulation.

\bibliographystyle{IEEEtran.bst}
\balance
\bibliography{IEEEabrv, references.bib}

\end{document}